# Integrating Triaxial IMU Sensors and Ensemble Learning for Effective Parkinson Disease Severity Classification

**Rehan Khan [1], Muhammad Ashraf [1], Sikandar Ali [2], Asad Ali Khan [3], Muhammad Junaid Asif [4], Nabeel Ahmad [5]**

[1] Department of Computer Sciences, COMSATS University Islamabad, Abbotabad 22060, Pakistan
[2] School of Advanced Studies, Center for Neuroscience, University of Camerino, Camerino 62032, Italy
[3] Department of Computer Sciences, Bahria University, Islamabad 44000, Pakistan
[4] Artificial Intelligence Technology Centre (AITeC), National Centre for Physics (NCP), Islamabad 44000, Pakistan
[5] Faculty of Computer Science and Information Technology, University of the Punjab, Lahore, Pakistan

**e-mail:** junaid.asif@ncp.edu.pk

***Abstract*—Parkinson disease (PD) is a progressive neurodegenerative disease that can have a significant impact on motor performance, resulting in the appearance of symptoms such as tremors, rigidity, postural instabilities and bradykinesia. Early and proper recognition of PD is closely linked to timely clinical treatment, disease management and the quality of life of the patient. In recent years, wearable sensor devices and AI have shown promise in developing non-invasive, data-informed disease detection techniques. In this paper, a comparative system is proposed based on artificial intelligence to detect Parkinson's disease by analyzing the motion and tremor data recorded by an inertial measurement unit (IMU). Data includes signals from the acceleration sensor and the gyroscope sensor which detect movement along the X, Y and Z axes. The signs and symptoms are useful for understanding subtle motor deficits seen with PD. The paper compared the effectiveness of several classification models such as Support Vector Machine (SVM), Logistic Regression (LR), K-Nearest Neighbors (KNN), Decision Tree (DT), Extreme Gradient Boosting (XGBoost), and Light Gradient Boosting Machine (LightGBM). Logistic Regression model achieved approximately 75% across all the assessment metrics and KNN model achieved approximately 90% performance. The support vector machine (SVM) achieved nearly 94% while the classifiers like Decision Tree and XGBoost achieved nearly 96% and overall classification efficacy, respectively. LightGBM model performs consistently at the best rank among all the evaluated methods, having Accuracy, Precision, Recall and F1-score of around 97%. The results demonstrate that the proposed machine learning approach is accurate and effective in predicting the severity of PD.**

## I. Introduction

Parkinson is a progressive nervous disorder, primarily linked to motor functions. It results from the progressive loss of nerve cells in the brain, causing a decrease in a chemical called dopamine that helps control movement. This has caused patients to experience tremors, slowness of movements (bradykinesia), rigidity of muscles, and an inability to balance. The disease is typically progressive and gradually reduces the patient's status and very serious in the daily activities. Particularly, PD is a health issue, which affects millions of people on earth, especially older people, and is one of the most serious health problems on earth given the significant long term effects on the quality of life [1], [2], [3].

Early diagnosis of Parkinson is essential for early therapy and prognosis. If the disease can be detected early, it can help the patient manage his/her symptoms more effectively and have the disease break out at a later stage. Besides, constant observation is critical since the symptoms change with time. The patient's tremor and movement profile can be used to gain a deeper understanding of the disease trajectory and tailor treatment accordingly. Thus, early diagnosis, as well as regular monitoring, is a critical way of managing the disease [4], [5].

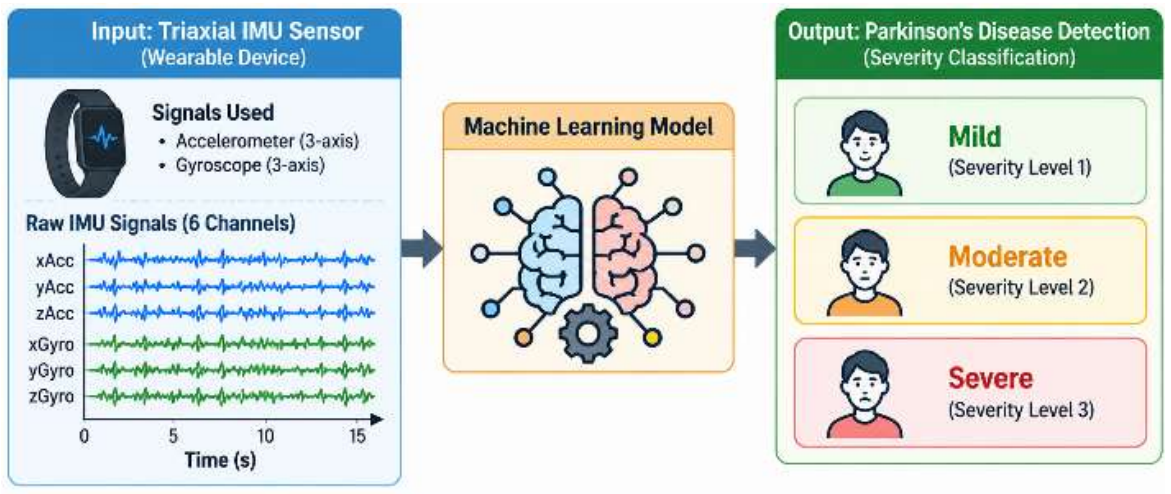

**Fig. 1.** Overview of the AI-powered Parkinson's disease severity classification framework using triaxial IMU sensor data

Parkinson disease is primarily diagnosed using conventional diagnostic techniques that are based on clinical observation and neurological examination [6]. These assessments are typically performed on a regular basis as part of a hospital visit and can only provide a limited overview of the patient's situation. In most cases, symptoms, such as tremors, may not be observed during an examination, making the evaluation difficult. Furthermore, these methods demand high-clinician expertise and subjectivity in making the diagnosis, which can lead to inconsistencies in diagnosis. Consequently, conventional approaches may not be accurate and frequent enough to provide an accurate disease monitoring [4].

Wearable technology has paved the way for new opportunities in continuous health monitoring for various assistive applications, including systems that provide haptic and auditory feedback to support individuals with visual impairments [7]. Wearable sensors are mini-research gadgets that can record real-time physiological and movement-related information [8]. In this category, Inertial Measurement Units (IMUs) are also widely used in body movement tracking as they contain accelerometers and gyroscopes which are essential for measuring acceleration, orientation and angular velocity [9]. In the healthcare sector, IMU sensors can also be used to identify abnormal movement patterns, which are associated with neurological diseases like Parkinson's disease. These devices enable continuous data collection, facilitating diagnosis and monitoring [10].

Simultaneously, AI and ML techniques have gained significant traction in the medical field due to their ability to handle vast amounts of data and identify intricate patterns [11]. ML models can be used to analyze motion and tremor data in the context of PD and detect subtle abnormalities which may be hard to detect manually [12]. AI solutions therefore have the potential to enhance detection of neurological diseases with greater accuracy and efficiency [13].

Although there is a lot of research on the detection and monitoring of PD, there are still many challenges. Numerous works have already been conducted with small-scale datasets limiting the models' generalizability and reliability when implemented in a clinical environment. Many methods are also not able to monitor the patient continuously and in real time, limiting their usefulness for long-term disease assessment. Furthermore, many of the existing methods do not fully exploit real-time motion and tremor data from wearable IMU sensors (accelerometers and gyroscopes) that can include relevant information on abnormal Parkinsonian movements. The above limitations highlight the critical need for more comprehensive data-driven and intelligent frameworks that utilize wearable sensing technologies along with sophisticated machine learning techniques to enhance the precision and reliability of PD severity detection and monitoring on the patient.

The primary aim of this study is to develop an accurate and reliable machine learning framework to detect the severity of Parkinson's disease using the motion and tremor data captured by the wearable IMU sensors. The study is designed to process signals from the triaxial accelerometer and the triaxial gyrometer which are collected along X, Y and Z axis to detect any irregularities in the movement and tremor pattern associated with Parkinson's disease. The accelerator measures linear movement and velocity change whereas the gyro sensor measures rotation and positioning. Additionally, several machine learning models are employed, leading to automation in disease severity classification, simplified early-stage detection, and facilitating continuous disease management of patients. The main contributions of this research paper are as follows:

- Comparative analysis of various machine learning based classification algorithms such as Logistic Regression (LR), K-Nearest Neighbours (KNN), Support Vector Machine (SVM), Decision Tree (DT), XGBoost and LightGBM are performed to determine the best and most reliable model for severity classification of Park
- The research emphasizes the potential of wearable IMU-based intelligent healthcare systems in promoting automated monitoring of PD and clinical decision-support.

Section II discusses the related work and previous research on the detection of Parkinson's disease using wearable sensor and machine learning methods. In Section III, the dataset, exploratory data analysis, and proposed methodology used for the classification of Parkinson's disease severity are described. The experimental results, performance evaluation and comparative discussion of the implemented machine learning models are given in Section IV. Lastly, Section V summarizes the paper and suggests future research directions and improvements.

## II. Related Work

### A. IMU-Based Detection of Gait Abnormalities in PD

Xie et al. [14] proposed a body-worn multisource gait monitoring system for quantitative gait analysis of PD disease. The system aimed to integrate multiple sensor modalities, including IMUs, force-sensitive sensors, and piezoelectric sensors, to collect data on various gait parameters, including plantar pressure, postural angles, and movement dynamics. The analysis was based on statistics and correlation of gait features obtained from the multisource gait set to evaluate gait abnormalities. The experimental results demonstrated the effectiveness of multisource sensor data for gait abnormality analysis as well as for use in monitoring and diagnosing Parkinson's disease. The study concluded that the features derived from the gyroscope data were more effective at differentiating between the On and Off medication states than the features derived from the accelerometer data. This indicates that gyroscope sensors can provide more detailed insights into the motor symptoms of Parkinson's disease. The investigators found that using IMU sensors to integrate accelerometer and gyroscope information could greatly enhance detection and monitoring of gait deviations. These sensor-based systems could enable automated monitoring of PD symptoms and aid physicians in early detection and more effective management of PD.

### B. Wearable Inertial Sensors for Monitoring of PD

A systematic review study examined the role of wearable inertial sensors in monitoring people with Parkinson's disease in their daily activities. The review examined 24 research papers published from 2010 to 2020 that utilized sensors like accelerometers, gyroscopes, and magnetometers to measure motor symptoms in patients. Most of these sensors were employed at home to record natural movements over extended durations. The studies were largely targeted at identifying gait changes, tremors, bradykinesia, dyskinesia and motor fluctuations. The results indicated that long-term motion data can be effectively captured by inertial sensors and that inertial sensors can be used to aid the identification of abnormal movements associated with Parkinson's disease. When utilizing ML techniques in tremor and motor symptom analysis, detection accuracies of over 90% have been reported in several studies. The review found wearable sensors could be very useful in helping to monitor PD continuously and help clinicians make more informed treatment decisions.

### C. Movement Disorder Detection using Wearable Inertial Sensors

Atri et al. [15] presented a system that monitors PD with wearable inertial sensors. The study was directed at detecting abnormal movement events and the symptoms of tremors, dyskinesia, and bradykinesia. The data were obtained from the wearable devices with accelerometers and gyroscopes fitted to the most affected upper limb of patients. The authors have designed a two-stage pipeline in which the first stage identifies normal or abnormal movement, and the second stage classifies movement disorder. The dataset was taken for both traditional machine learning models as well as a deep learning model ***(CNN-RNN based architecture)***. The outcome revealed that deep learning method achieved high performance where the accuracy in detecting movement disorder was approximately 93% and the accuracy in symptom classification was approximately 91.5%. The study found that integrating wearable sensors with machine learning can be a useful tool for tracking Parkinson's symptoms, and that wearable technology could be used for real-time health monitoring with devices like smartwatches and fitness bands.

### D. IMU-Based Kinetic Tremor Detection in PD

Jung et al. [16] presented an approach to detect kinetic tremors in Parkinson's patients from IMU sensing data. The sensors attached to the wrists of 30 patients were used during drawing tasks in this study, including spiral drawing and straight lines, which are frequently used in the clinical examination of tremor. Motion data such as 3-axis accelerometer, gyroscope, and magnetometer signals were acquired via the sensors at 100 Hz. The data collected was processed to detect tremor episodes and determine the amplitude of the tremor. Given this, the time domain sensor data were transformed to Continuous Wavelet Transform (CWT) images to allow an analysis of both time and frequency information. The analysis showed that patients with tremor symptoms had a predominant tremor frequency near 8 Hz. The study also calculated amplitude of the tremor by differentiating the accelerometer data, producing velocity data. Results demonstrated that this method can easily identify tremor patterns during movement activities. The researchers found that IMU wearable devices, such as smartwatches, have great potential to be a non-invasive monitoring and measurement tool for tremors in PD patients.

### E. Movement Disorder Detection Using Wearable Inertial Sensors

A recent study [17] proposed a system for monitoring symptoms of Parkinson's disease in real-time using a combination of wearable sensors and inertial data from smartwatches and fitness tracking bands. In this system, the data from the triaxial accelerometer of 28 Parkinson patients are used and a two-classifier system is proposed to detect abnormal movements and classify the patient as tremor or dyskinesia or bradykinesia. ***Random Forest, HARDenseRNN*** and conventional machine learning and deep learning techniques were evaluated. The results of the experiments were good, with the deep learning model achieving 93.03% recall for the movement disorder detection and 91.54% recall for the classification of symptoms. The study highlighted the promising opportunity of employing smart algorithms and ML techniques to monitor and evaluate PD remotely from real-time.

## III. MATERIALS AND METHODS

### A. Dataset Details

The dataset is available on Kaggle [18], which consists of data recorded from multiple Inertial Measurement Unit (IMU) sensors on an accelerometer and gyroscope sensor. The dataset consists of data collected from various samples, each of which is assigned a label corresponding to one of three severity classes ***(as specified in Table 1)*** to indicate the stage of symptoms of PD. For both these types of sensing, the motion value is recorded digitally in three spatial directions ***(x, y and z).*** The accelerometer readings are linear, while the gyroscope readings are angular. The multi-modal motion data contains much information regarding tremor, abnormal gait and motor issues related to Parkinson's disease. The characteristics of dataset are as shown in ***Table 2.*** The data was thoroughly cleaned and pre-processed for which no further process was needed. It is balanced, meaning that there are equal numbers of samples on each label of it (1, 2 and 3), making machine learning models more likely to be accurate.

**Table 1.** Parkinson Disease Severity Labels

| ***Class Label*** | ***Severity Level*** | ***Description*** |
|---|---|---|
| 1 | Low | Mild Parkinsonian symptoms |
| 2 | Moderate | Intermediate Parkinsonian symptoms |
| 3 | High | Severe Parkinsonian symptoms |

**Table 2.** Parkinson Disease Severity Labels

| ***Parameter*** | ***Description*** |
|---|---|
| Data Type | Parkinson's IMU Tremor Severity Dataset [18] |
| Data Collection site | Liaquat University Hospital, Hyderabad, PK |
| Sensors Used | Accelerometer and Gyroscope |
| Data Acquisition Device | Arduino Nano 33 BLE Sense |
| No. of Channels | 6 Nos. |
| No. of Axis | 3 ***(x,y,z)*** |
| No. of Classes | 3 |
| No. of Samples | 4500 Nos. |
| No. of Pateints | 15 Nos. |
| Recording Condition | Resting hand Tremor |
| Recording Duration | 100 Seconds per Patient |
| Data Split | Train:Test = 80:20 (Patient Level) |

### B. Proposed Method

The proposed approach is to detect and classify the intensity of Parkinson's disease by using motion and tremor signals captured from the wearable triaxial IMU sensors (accelerometers and gyroscopes) ***as shown in Fig. 2***. These sensors can give detailed information on the movements while measuring the angular velocity in the X, Y and Z axes and linear acceleration in X and Y axes, indicating tremor and abnormality of movements, which are associated with Parkinson's disease. The raw IMU sensor data is initially analyzed for Exploratory Data Analysis (EDA) for the exploration of distributions in the signals, their relations, and their characteristics of movement. The proposed framework does not require a lot of preprocessing and handcrafted signal transformation of the raw sensor feature matrix, which is a matrix of values from the accelerometer and the gyroscope, as in the conventional approaches. This leads to retaining important movements and eventually reduces computational complexity. The feature matrix is then fed into various machine learning classifiers—namely ***Logistic Regression (LR), K-Nearest Neighbors (KNN), Support Vector Machine (SVM), Decision Tree (DT), XGBoost and LightGBM***. The models are trained to recognize discriminative patterns from the triaxial motion signals and to distinguish severity of Parkinson's into three stages: mild, moderate and severe.

The proposed methodology has six major steps: Data Acquisition, Exploratory Data Analysis and Feature Matrix Definition, Model Training, Model Performance Evaluation, Deployment and Clinical Support. During training, the training data is divided into training data and testing data to evaluate the model's ability to generalize. The proposed method utilizes accelerometer and gyroscope motion data to accurately analyze tremor characteristic and motion to provide high reliability and accuracy prediction of the severity of Parkinson's disease. Furthermore, the architecture is scalable, interpretable, and suitable for intelligent healthcare application, as that the clinicians and researchers can consider the system as an assisting component of the automated monitoring and clinical decision-making process of a Parkinson's disease.

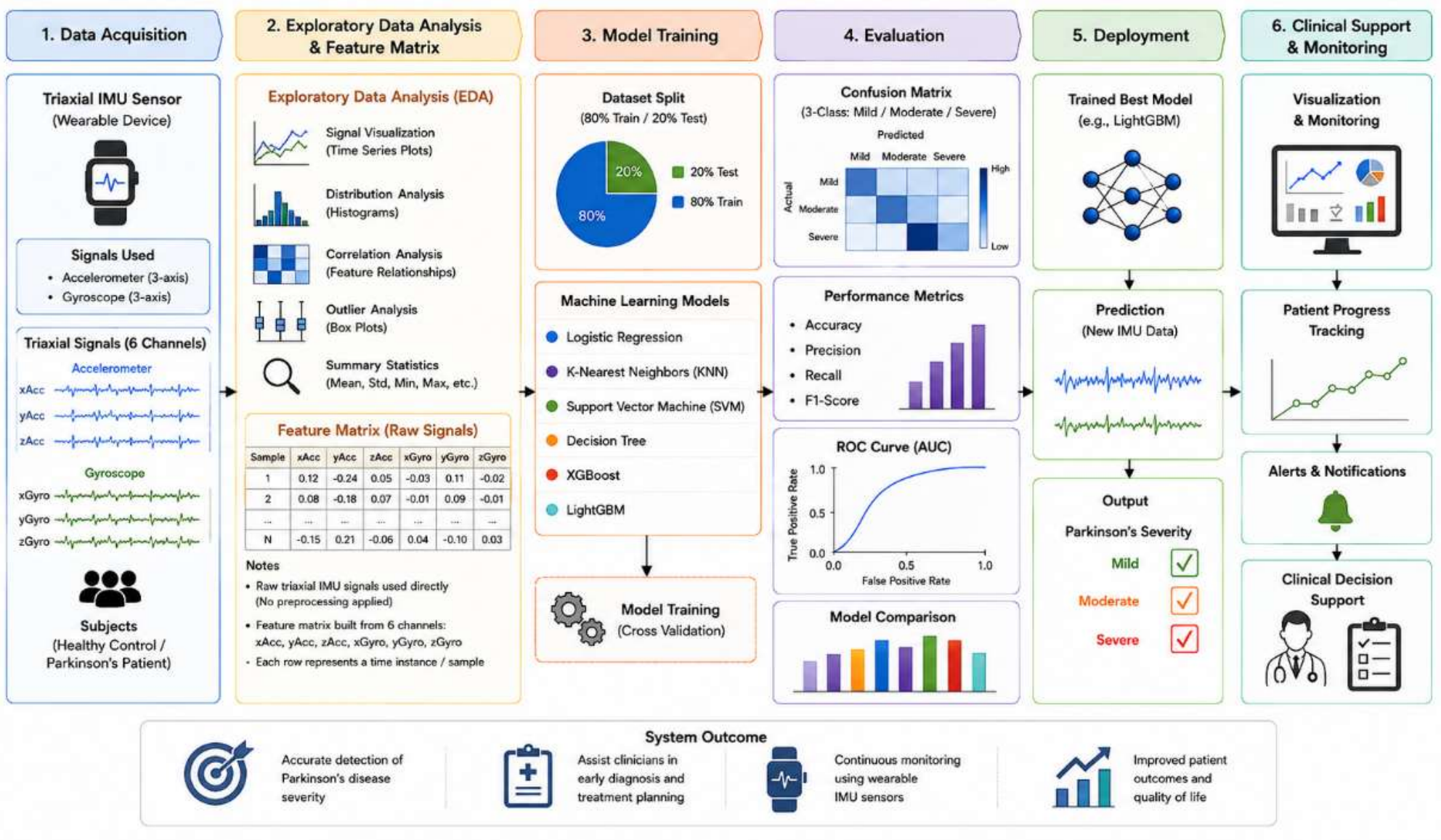

**Fig. 2.** Proposed framework for AI-powered Parkinson's disease severity analysis using triaxial IMU sensor data and machine learning techniques

## *C. Evaluation Metrics*

Evaluation of the proposed framework is done by using standard evaluation metrics such as accuracy, precision, recall and F1-score. The metrics offer a holistic evaluation of the models' performance at classifying tremors in Parkinson's patients. The accuracy in determining the percentage of correct classification out of the total number of classifications is correlated with the correctness of the classification model. Precision is used to assess the accuracy of the positive prediction: it is the percentage of observed positive cases compared with the number of positive cases that is predicted. Recall/Sensitivity means the ability of the model to identify all the true positives in the dataset [19], [20] [21] [22] [23]. The F1 score is important when assessing the robustness of the classification, because it considers the accuracy of the classifications as well as their recall, giving a balanced assessment. These evaluation parameters provide comprehensive information about the effectiveness, reliability, and robustness of the ML in detecting and classifying the severity of PD.

$$\text{Accuracy} = \frac{\text{No.of Correct Predictions}}{\text{Total number of predictions}} \quad \textbf{(I)}$$

$$\text{Precision} = \frac{\text{True Positives}}{\text{True Positives+Fal}\quad\text{Positives}} \quad \textbf{(II)}$$

$$\text{Recall} = \frac{\text{True Positives}}{\text{True Positives+False Negatives}} \quad \textbf{(III)}$$

$$\text{F1} - \text{Score} = 2\text{x}\,\frac{(\text{Precision x Recall})}{(\text{Precison+Reca}\quad)} \quad \textbf{(IV)}$$

# IV. Results

This section outlines the performance of the proposed machine learning model and presents the experimental results in determining the severity of Parkinson's disease based on the data retrieved from a combination of the acceleration and gyroscope signals from the triaxial IMU sensor. The confusion matrix and graphical visualizations were also included for comparative analysis and to help understand the classification ability of each model.

## *A. Performance with Logistic Regression*

The Logistic Regression model achieved moderate performance with a relatively high classification accuracy for both severity and level classes in Parkinson's disease (***as shown in Fig. 3 (a)***). The model achieved an overall accuracy of ***75%.*** The classification performance is not different between any severity class since after the macro averaging, they were also 75% for Precision, Recall and F1-score. The results revealed that Logistic Regression could extract most of these from the sensor data collected to measure fundamental tremors from intrusive sensors installed at IMU, giving a satisfactory result. Furthermore, F1 score has been shown to be aware of the balance between precision and recall. Logistic Regression is a simple, easily interpretable and fast to compute model, which is why it's a good model to use for comparison with more complex ensemble methods.

## *B. Performance with KNN*

The performance of K-Nearest Neighbors (KNN) was significantly better than Logistic Regression and is presented graphically in ***Fig. 3 (b)***. This model was able to detect a pattern of tremor severity from sensor data, with an accuracy of approximately ***90%***, precision, recall and F1-score, showing successful identification of the patterns. All the Metric values are balanced, which suggests balanced classification performance in all severity classes. The results show that KNN can effectively capture the local relationship between features and similarity within the triaxial motion signals.

## *C. Performance with SVM*

For the Parkinson's disease severity detection task, Support Vector Machine (SVM) performed well in classification, as presented in ***Fig. 3 (c)***. From the above results, it can be concluded that the accuracy, precision and recall for each class are around ***94%***, which shows that the results are very close to the true results with a balanced prediction in all classes. Based upon the accelerometer data and the gyroscopic data, SVM outperforms because it can create optimal decision boundaries and model complex motion patterns of the oscillating device, which is related to the tremors.

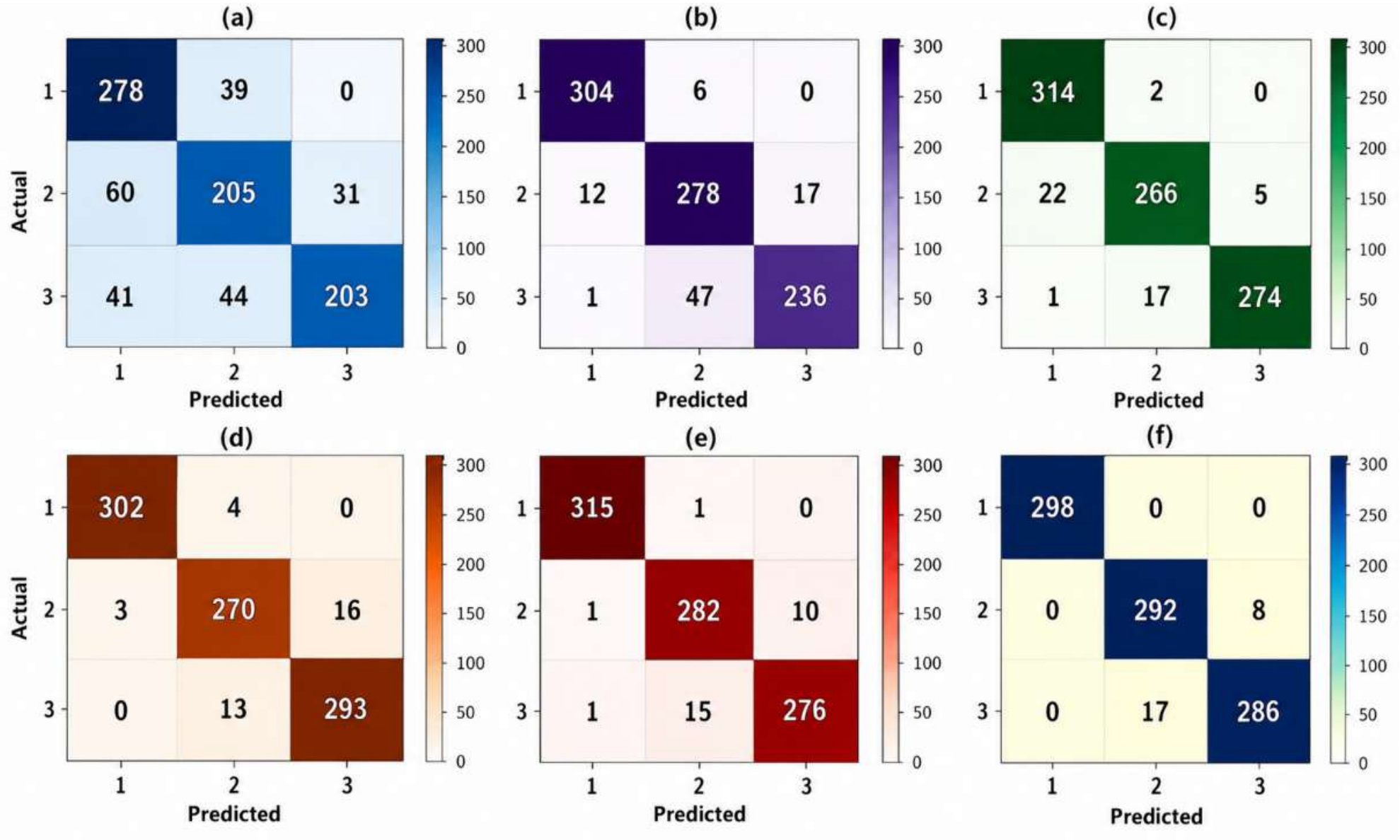

**Fig. 3.** Confusion matrices of the evaluated machine learning models for Parkinson's disease severity classification using IMU sensor data: (a) Decision Tree, (b) Random Forest, (c) Support Vector Machine (SVM), (d) K-Nearest Neighbors (KNN), (e) Logistic Regression, and (f) Naïve Bayes. The diagonal values represent correctly classified samples, while off-diagonal values indicate misclassifications among the three severity classes.

### *D. Performance with Decision Tree*

The Decision Tree resulted in a very good performance when it came to the classification of the severity of the PD as demonstrated in ***Fig. 3(d)***. The model obtained about ***96%*** Accuracy, Precision, Recall as well as F1-score, i.e. the model can learn the discriminating movement patterns in the IMU sensor data. The results indicate that the Decision Tree model is a good way to represent the interactions of features when they are nonlinear and it has good classification performance without losing interpretability.

### *E. Performance with XGBoost*

The XGBoost classifier has also shown excellent predictive performance as illustrated in ***Fig. 3 (e)***. The model showed extremely high accuracy, precision, recall and F1-score values in the classification category which is approximately 96% each. The high results of XGBoost indicate the power of ensemble accelerating strategies during wider learning covered unseen associations and multifaceted tremor attributes using triaxial signals of IMU sensors.

### *F. Performance with LGBM*

As seen in ***Fig. 3 (f)***, LightGBM Classifier is performing better among all the classifiers. The model has high Accuracy, Precision, Recall and F1 score about 97%, thus demonstrating its potential to be better for classifying the severity of Parkinson disease. The efficiency of the LightGBM to handle complex motion anomalies and the accelerometer and gyroscope measurement characteristics of the tremors are highlighted from the balanced high metrics. The results indicated that LightGBM is the best and the most accurate classifier among the classifiers used and compared.

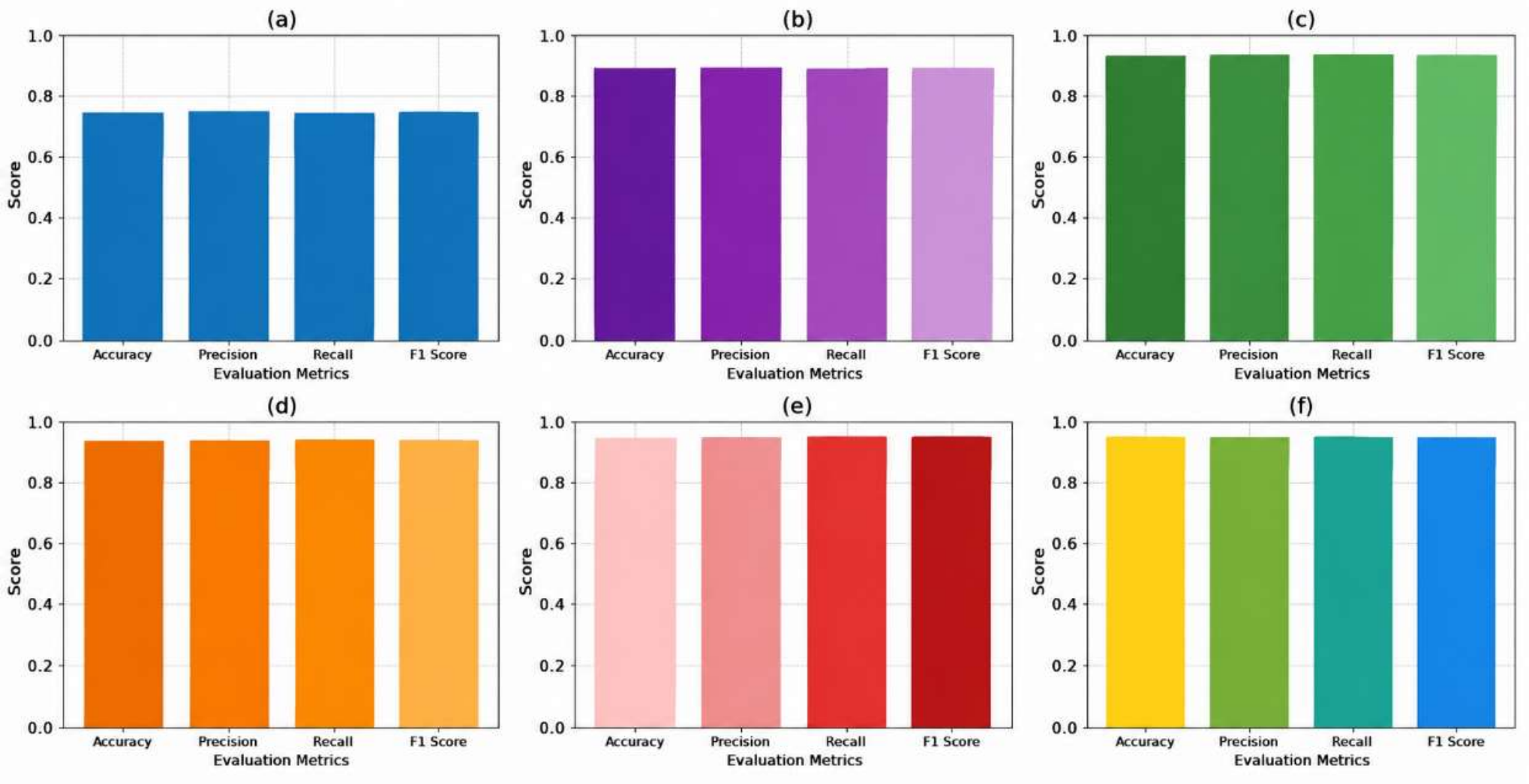

**Fig. 4.** Performance evaluation metrics of the implemented machine learning models for Parkinson's disease severity classification using IMU sensor data: (a) Logistic Regression, (b) K-Nearest Neighbors (K-NN), (c) Support Vector Machine (SVM), (d) Decision Tree, (e) XGBoost, and (f) LightGBM. The bar plots illustrate the comparative performance in terms of Accuracy, Precision, Recall, and F1-Score for each classification

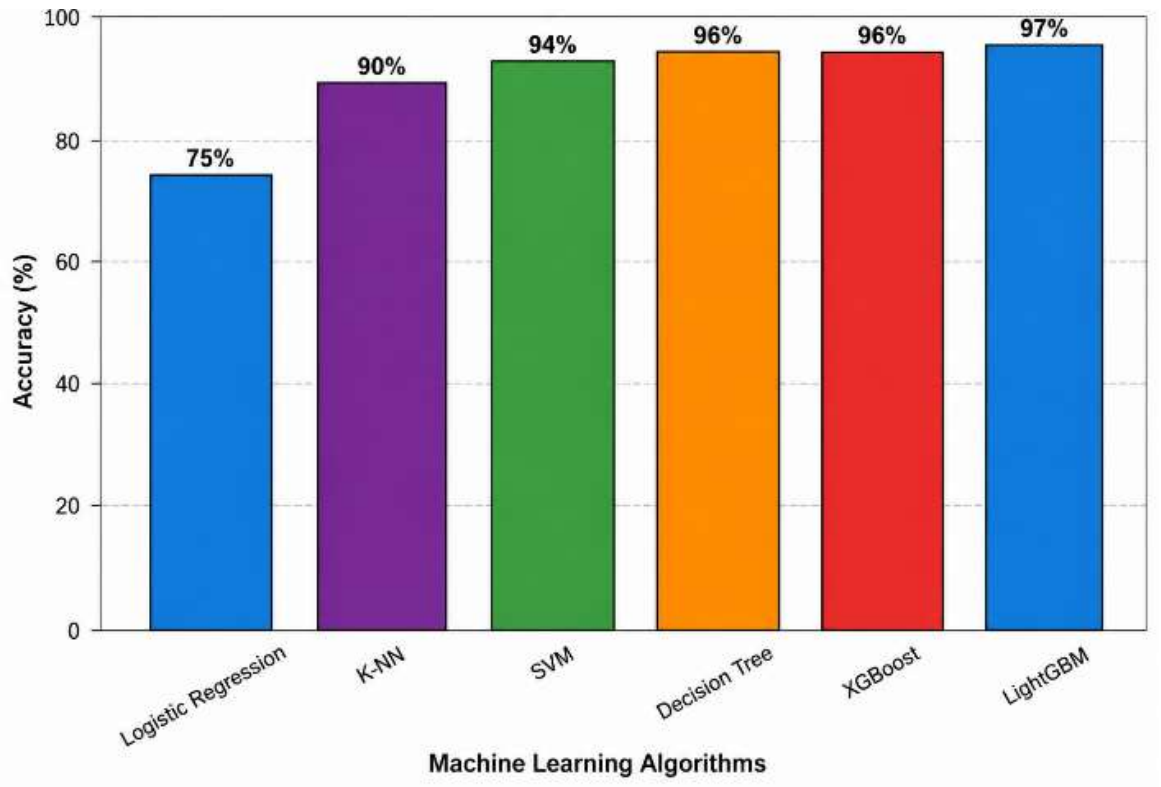

**Fig. 5.** Comparative analysis of machine learning algorithms for Parkinson's disease detection using IMU sensor data.

## V. CONCLUSIONS

This comparative study of the six machine learning models implemented shows a wide range of potential to classify the severity of Parkinson's Disease based on triaxial IMU sensor data. The best performance was achieved by ***Logistic Regression (around 75% Accuracy, Precision, Recall, and F1-score)*** which serves as a baseline model for comparison of the other results. The model was found to offer good balance of predictions, but since it was linear, it was not sufficiently able to efficiently represent the complex nonlinear tremor and movement patterns in the sensor signals. By using similarity-based learning, the K-Nearest Neighbors (KNN) classifier has achieved a performance of around 90% on all measures of evaluation, showing that similarity-based learning is able to classify items better. Similarly, Support Vector machine (SVM) model had highest predictive accuracy of approximately 94%, which is good in detecting high dimensions of IMU sensor data and in capturing complicated decision boundaries. Ensemble-based approaches were able to further enhance performance. The Decision Tree classifier and XGBoost classifier obtained around 96% of all evaluation parameters indicating their ability to attain nonlinear relationships of features and the features motion not seen in plain predictors but latent in the Parkinsonian tremors. In this experiment, the LightGBM classifier had the best overall performance, with a high F1-score of around 97%, Precision, Recall, and Accuracy and was thus considered the most reliable and efficient model of the severity of Parkinson's disease in this case study. The macro averages of Accuracy, Precision, Recall and F1-score for each model are stable, which indicates good performance of the models when trained for macro averages. Moreover, the performance clearly depicted in the graphical analysis, reveals that the advanced ensemble learning techniques are better than introductory machine learning methods, where LightGBM and XGBoost method outperforms the conventional method. Overall, the outcome of the experiment suggests that gradient boosting models could be highly beneficial in the interpretation of complex tremor and movement patterns in wearable IMU sensor data in an analysis of Parkinson's disease. It is a promising model of intelligent systems for monitoring and clinical decision making in PD.

## REFERENCES

[1] N. Saini *et al.*, "Motor and non-motor symptoms, drugs, and their mode of action in Parkinson's disease (PD): a review," *Medicinal Chemistry Research*, vol. 33, no. 4, pp. 580--599, 2024.

[2] R. Balestrino and A. H. V Schapira, "Parkinson disease," *Eur. J. Neurol.*, vol. 27, no. 1, pp. 27--42, 2020.

[3] B. R. Bloem, M. S. Okun, and C. Klein, "Parkinson's disease," *The Lancet*, vol. 397, no. 10291, pp. 2284--2303, 2021.

[4] W. Wang, J. Lee, F. Harrou, and Y. Sun, "Early detection of Parkinson's disease using deep learning and machine learning," *IEEE access*, vol. 8, pp. 147635--147646, 2020.

[5] N. Ghosh, K. Sinha, and P. C. Sil, "A review on the new age methodologies for early detection of Alzheimer's and Parkinson's disease," *Basic Clin. Pharmacol. Toxicol.*, vol. 134, no. 5, pp. 602--613, 2024.

[6] M. J. Armstrong and M. S. Okun, "Diagnosis and treatment of Parkinson disease: a review," *JAMA*, vol. 323, no. 6, pp. 548--560, 2020.

[7] S. T. H. Rizvi, M. J. Asif, and H. Ashfaq, "Visual impairment aid using haptic and sound feedback," in *2017 International Conference on Communication, Computing and Digital Systems (C-CODE)*, 2017, pp. 175–178.

[8] E. P. Adeghe, C. A. Okolo, and O. T. Ojeyinka, "A review of wearable technology in healthcare: Monitoring patient health and enhancing outcomes," *OARJ of Multidisciplinary Studies*, vol. 7, no. 01, pp. 142--148, 2024.

[9] G. Maniatis, "On the use of IMU (inertial measurement unit) sensors in geomorphology," *Earth Surf. Process. Landf.*, vol. 46, no. 11, pp. 2136--2140, 2021.

[10] F. Bo *et al.*, "IMU-based monitoring for assistive diagnosis and management of IoHT: a review," 2022.

[11] L. Rubinger, A. Gazendam, S. Ekhtiari, and M. Bhandari, "Machine learning and artificial intelligence in research and healthcare," *Injury*, vol. 54, pp. S69--S73, 2023.

[12] E. Abdulhay, N. Arunkumar, K. Narasimhan, E. Vellaiappan, and V. Venkatraman, "Gait and tremor investigation using machine learning techniques for the diagnosis of Parkinson disease," *Future Generation Computer Systems*, vol. 83, pp. 366--373, 2018.

[13] G. Battineni, G. G. Sagaro, N. Chinatalapudi, and F. Amenta, "Applications of machine learning predictive models in the chronic disease diagnosis," *J. Pers. Med.*, vol. 10, no. 2, p. 21, 2020.

[14] J. Xie *et al.*, "Wearable multisource quantitative gait analysis of Parkinson's diseases," *Comput. Biol. Med.*, vol. 164, p. 107270, 2023.

[15] R. Atri *et al.*, "Deep Learning for Daily Monitoring of Parkinson's Disease Outside the Clinic Using Wearable Sensors," *Sensors*, vol. 22, no. 18, p. 6831, Sep. 2022, doi: 10.3390/s22186831.

[16] W. Jung *et al.*, "IMU Sensing Data-Based Kinetic Tremor Detection in Parkinson's Disease Patients," in *Proceedings of the 20th ACM Conference on Embedded Networked Sensor Systems*, New York, NY, USA: ACM, Nov. 2022, pp. 772–773. doi: 10.1145/3560905.3568052.

[17] U. Khan, Q. Riaz, M. Hussain, M. Zeeshan, and B. Krüger, "Towards Effective Parkinson's Monitoring: Movement Disorder Detection and Symptom Identification Using Wearable Inertial Sensors," *Algorithms*, vol. 18, no. 4, p. 203, Apr. 2025, doi: 10.3390/a18040203.

[18] "{P}arkinson's {I}{M}{U} {T}remor {S}everity {D}ataset --- kaggle.com."

[19] M. J. Asif, "Crowd Scene Analysis using Deep Learning Techniques," University of Central Punjab (UCP) Lahore, Pakistan, 2024.

[20] M. J. Asif *et al.*, "Exploring Convolutional Neural Networks for Rice Grain Classification: An Explainable AI Approach," in *2025 International Conference on Frontiers of Information Technology (FIT)*, 2025.

[21] M. J. Asif, M. S. Rafaqat, U. Nazakat, U. Khan, and R. F. Ahmad, "Towards Automated Solar Panel Integrity: Hybrid Deep Feature Extraction for Advanced Surface Defect Identification," 2026.

[22] M. J. Asif, M. Asad, and S. Imran, "Crowd Scene Analysis: Crowd Counting using MCNN based on Self-Supervised training with Attention Mechanism," in *2023 25th International Multitopic Conference (INMIC)*, 2023, pp. 1–7.

[23] H. Khalid, S. Saqib, M. J. Asif, and D. A. Dewi, "Strategic Customer Segmentation: Harnessing Machine Learning For Retaining Satisfied Customers," *Lahore Garrison University Research Journal of Computer Science and Information Technology*, vol. 8, no. 2, 2024.